\documentclass[conference]{IEEEtran}
\IEEEoverridecommandlockouts
\usepackage{cite}
\usepackage{amsmath,amssymb,amsfonts}
\usepackage{algorithmic}
\usepackage{graphicx}
\usepackage{textcomp}
\usepackage{xcolor}
\usepackage{booktabs}
\usepackage{url}
\usepackage{hyperref}

\def\BibTeX{{\rm B\kern-.05em{\sc i\kern-.025em b}\kern-.08em
		T\kern-.1667em\lower.7ex\hbox{E}\kern-.125emX}}
\begin{document}
	
	\newcommand{\Eq}[1]{Equation~(\ref{#1})}
	\newcommand{\Sect}[1]{Section~\ref{#1}}
	\newcommand{\Sects}[1]{Sections~\ref{#1}}
	\newcommand{\Fig}[1]{Figure~\ref{#1}}
	\newcommand{\Table}[1]{Table~\ref{#1}}
	\newcommand{\iec}{i.\,e.,\,}
	\newcommand{\ie}{i.\,e.\,}
	\newcommand{\egc}{e.\,g.,\,}
	\newcommand{\eg}{e.\,g.\,}

	\title{Audio-Conditioned U-Net for\\ Position Estimation in Full Sheet Images}
	
	\author{\IEEEauthorblockN{Florian Henkel}
		\IEEEauthorblockA{\textit{Institute of Computational Perception} \\
			\textit{Johannes Kepler University}\\
			Linz, Austria \\
			florian.henkel@jku.at}
		\and
		\IEEEauthorblockN{Rainer Kelz}
		\IEEEauthorblockA{\textit{Austrian Research Institute for} \\
			\textit{Artificial Intelligence}\\
			Vienna, Austria \\
			rainer.kelz@ofai.at}
		\and
		\IEEEauthorblockN{Gerhard Widmer}
		\IEEEauthorblockA{\textit{Institute of Computational Perception} \\
			\textit{Johannes Kepler University}\\
			Linz, Austria \\
			gerhard.widmer@jku.at}
	}
	
	\maketitle
	
	\begin{abstract}
		The goal of score following is to track a musical performance, usually in the form of audio, in a corresponding score representation. Established methods mainly rely on computer-readable scores in the form of MIDI or MusicXML and achieve robust and reliable tracking results. 
		
		Recently, multimodal deep learning methods have been used to follow along musical performances in raw sheet images. Among the current limits of these systems is that they require a non trivial amount of preprocessing steps that unravel the raw sheet image into a single long system of staves.
		
		The current work is an attempt at removing this particular limitation. We propose an architecture capable of estimating matching score positions directly within entire unprocessed sheet images.
		We argue that this is a necessary first step towards a fully integrated score following system that does not rely on any preprocessing steps such as optical music recognition.

	\end{abstract}
	
	
	\section{Introduction}
	A large body of work on score following requires the use of a computer-readable representation of the score, e.g., MusicXML or MIDI \cite{DixonW05_MATCH_ISMIR, Arzt16_MusicTracking_PhD, OrioLS03_scorefollowing_NIME, Cont06_ScoreFollowingViaNmfAndHMM_ICASSP,
		SchwarzOS04_scoreFollowing_ICMC, NakamuraCCOS15_ScoreFollowingSemiHMM_ISMIR}. Such representations can either be manually created, which is tedious and time consuming, or extracted using optical music recognition (OMR). Recently, score following has also been demonstrated with raw sheet images, using multimodal deep (reinforcement) learning \cite{DorferAW16_ScoreFollowDNN_ISMIR,DorferHW18_ScoreFollowingAudioSheet_ISMIR}. However, these latter approaches still rely on several preprocessing steps -- in particular, the score must be `unrolled' into a single long staff: consecutive staves need to be detected on the score sheet, cut out, and presented to the score following system in 
	sequence.
	
	In this work, we propose an architecture that can directly predict which parts in a complete, unprocessed score page match a given audio excerpt.
	Our system is based on a U-Net \cite{RonnebergerFB15_UNET_MICCAI} for musical symbol detection \cite{HajicDWP18_OMRUNET_ISMIR} and uses Feature-wise Linear Modulation (FiLM) layers \cite{PerezSDVDC18_FILM_AAAI}. A similar architecture was recently used for the task of music source separation \cite{MeseguerP19_CUNET_ISMIR}.
	As a proof of concept we test our model on simple monophonic music from the Nottingham dataset \cite{BoulangerBV12_ModelingTemporalDependencies_ICML}.
	While in its current state the system is not a full score follower yet because it ignores temporal context constraints, the capability to directly process full score sheet images is arguably a necessary first step.
	
	The remainder of this paper is structured as follows.
	In~\Sect{sec:architecture}, we introduce our proposed architecture and its underlying concepts. In~\Sect{sec:experiments}, we conduct several experiments to compare different architecture choices and to show that our system is indeed able to predict score positions in sheet images.\footnote{To get a better impression of what our network does we provide videos: \url{https://github.com/CPJKU/audio_conditioned_unet}}
	Finally, in~\Sect{sec:conclusion}, we summarize our work and provide an outlook on how to adapt the architecture for more complex scores and how to incorporate temporal context for a full score following setting.
	
	\section{Audio-Conditioned U-Net for Position Estimation in Sheet Images}\label{sec:architecture}
	
	\begin{figure*}[t]
		\centering
		\includegraphics[width=0.80\textwidth]{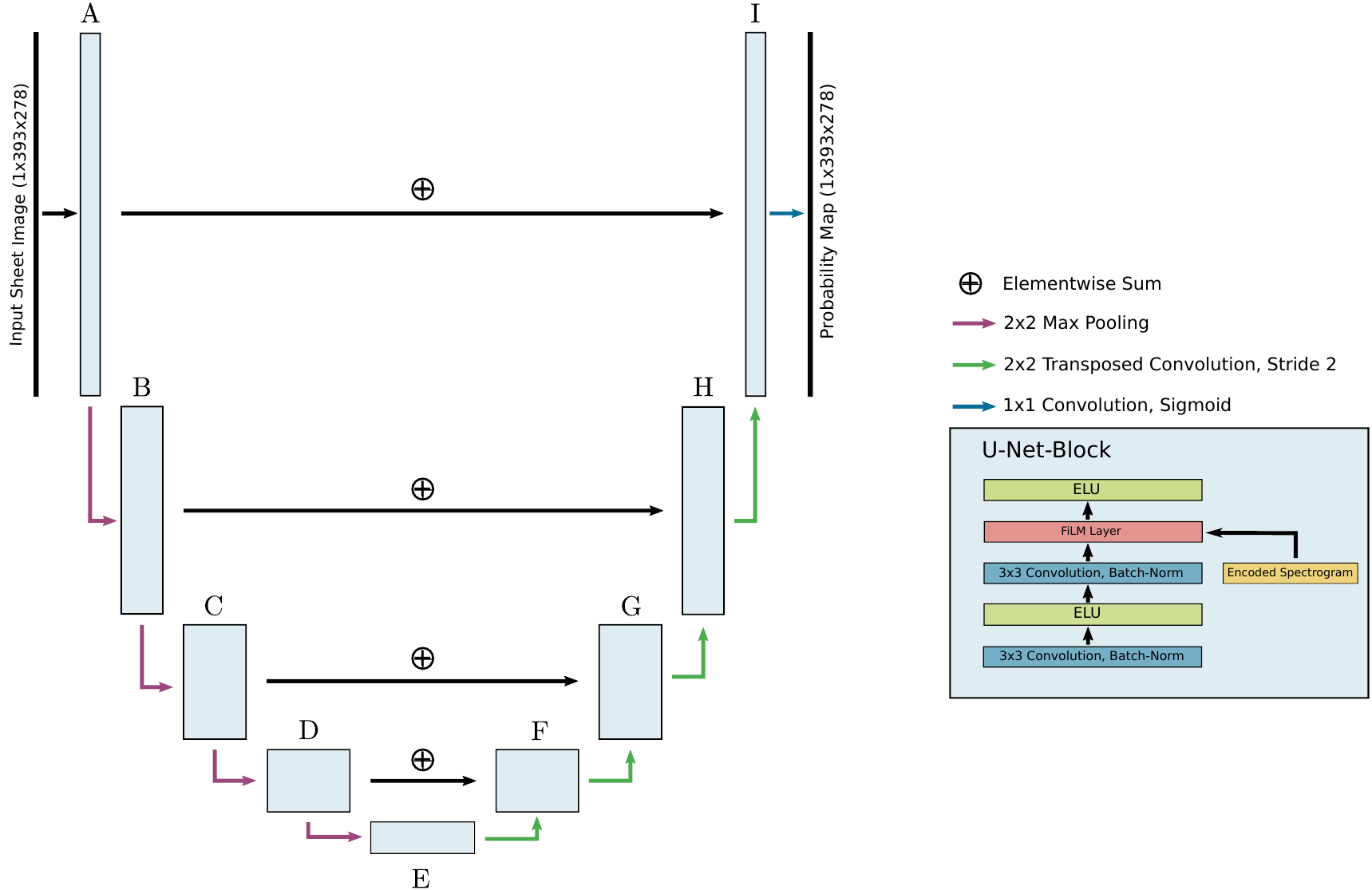}
		\caption{Audio-Conditioned U-Net architecture. Each block (A-I) consists of two convolutional layers followed by batch normalization  and the ELU activation function. 
		The FiLM layer is placed before the last activation function. The audio spectrogram encoding used for conditioning is given by the output of the network shown in \Table{tab:spec_net}. Each symmetric block has the same number of filters starting with 8 in block A and increasing with depth to 128 in block E.}
		\label{fig:architecture}
	\end{figure*}
	
	U-Nets are fully-convolutional neural networks that were introduced for the task of medical image segmentation \cite{RonnebergerFB15_UNET_MICCAI}. They can be used to segment an image into different parts, \egc by classifying each pixel into either foreground or background.
	In \cite{HajicDWP18_OMRUNET_ISMIR}, Haji{\v{c}} et al. use U-Nets for detecting musical symbols in sheet images. We adapt this architecture to predict positions in sheet images that correspond to a given audio excerpt, \iec we segment the sheet image into regions that match the audio snippet, and regions that do not. To this end, we include \textit{Feature-wise Linear Modulation (FiLM)} layers \cite{PerezSDVDC18_FILM_AAAI} as a conditioning mechanism in the U-Net architecture, as shown in \Fig{fig:architecture}. Each FiLM layer applies a simple affine transformation to the feature maps it is connected to, conditioned on an external input.
	In our case, the conditioning input is an encoded representation $\mathbf{z}$ of the audio excerpt, which is created by another neural network depicted in \Table{tab:spec_net}. The FiLM layer is defined as:
	\begin{equation}
	\text{FiLM}(\mathbf{x}) = \gamma(\mathbf{z}) \cdot \mathbf{x} + \beta(\mathbf{z}),
	\end{equation}{}
	with $\gamma(\cdot)$ and $\beta(\cdot)$ being learned functions. Each scalar output component $\gamma_k(\cdot)$, $\beta_k(\cdot)$ scales and shifts the feature map $\mathbf{x}_k$, where $\mathbf{x}_k$ is the $k$-th output of a convolutional layer  with $K$ number of filters, after batch normalization is applied. 
	
	Previous work combining sheet images and audio mainly relies on an embedding space for these two modalities which is achieved by combining the representation of two separate networks \cite{DorferHAFW18_MSMD_TISMIR, DorferHW18_ScoreFollowingAudioSheet_ISMIR, DorferAW16_ScoreFollowDNN_ISMIR}. FiLM layers on the other hand directly interfere in the learned representation of the sheet image by modulating its features, which helps the network to focus only on those parts that are required for a correct prediction.
	Additionally, their inherent structure allows us to use a fully convolutional architecture for the sheet image, \iec the network can process scores of arbitrary resolution, which will be useful for exploring generalization capabilities in future work.
	
	The U-Net has a basic encoder-decoder structure, and consists of four down-sampling blocks (A-D), four up-sampling blocks (F-I) and a bottleneck block (E). Down-sampling is done by $2\times 2$ max pooling and for up-sampling we use transposed convolutions with a filter size of $2 \times 2$.
	After each down-sampling step, the number of filters is doubled (starting at 8), whereas each up-sampling step halves the number of filters, \iec block A has 8 filters, E has 128 and I has again 8.
	
	Each block consists of two convolutional layers followed by batch normalization \cite{IoffeS15_BatchNorm_arxiv} and the ELU activation function \cite{ClevertUH15_ELU_ICLR}. The FiLM layer is placed after the last batch normalization layer and before the activation function. Adhering closely to \cite{HajicDWP18_OMRUNET_ISMIR}, we add residual connections in the form of an element-wise sum between symmetric building blocks, as depicted in \Fig{fig:architecture}.
	The output layer consists of a $1\times1$ convolution followed by the sigmoid activation function, yielding a per-pixel pseudo probability map that highlights regions in the score corresponding to a given audio excerpt.

	\section{Experiments}\label{sec:experiments}
	In the following, we describe data and ground truth required in our experiments, introduce the experimental setup, and discuss the performance of different architecture choices.
	
	\begin{table}
		\small
		\centering
		\begin{tabular}{c}
			Audio (Spectrogram)  $78 \times40$\\
			\toprule
			Conv 16x3x3 - stride-1 - BN - ELU\\
			Conv 16x3x3 - stride-1 - BN - ELU\\
			
			Conv 32x3x3 - stride-2 - BN - ELU\\
			Conv 32x3x3 - stride-1 - BN - ELU\\
			
			Conv 64x3x3 - stride-2 - BN - ELU\\
			Conv 96x3x3 - stride-2 - BN - ELU\\
			
			Conv 96x1x1 - stride-1 - BN - ELU\\
			
			Linear 128 - BN - ELU \\
			
			\midrule
		\end{tabular}
		\caption{The Spectrogram Encoder. We use batch normalization (BN)\cite{IoffeS15_BatchNorm_arxiv} and the ELU activation function \cite{ClevertUH15_ELU_ICLR}. The network structure resembles the one used in \cite{DorferHW18_ScoreFollowingAudioSheet_ISMIR}.}
		\label{tab:spec_net}
	\end{table}

	\subsection{Data}
	We use a subset of the \emph{Nottingham} dataset, comprising 274 monophonic melodies of folk music, partitioned into 172 training, 60 validation and 42 test pieces \cite{BoulangerBV12_ModelingTemporalDependencies_ICML}. The sheet music is created by automatically typesetting the MIDI scores with Lilypond,\footnote{\url{http://lilypond.org/}} and the audio is synthesized using a piano sound font.
	The rendered score images have a resolution of $1181\times835$ pixels, and are downscaled by a factor of 3 to $393\times278$ pixels, to be used as input to the convolutional neural network.
	
\begin{figure*}[t]
		\centering
		\includegraphics[width=0.75\textwidth]{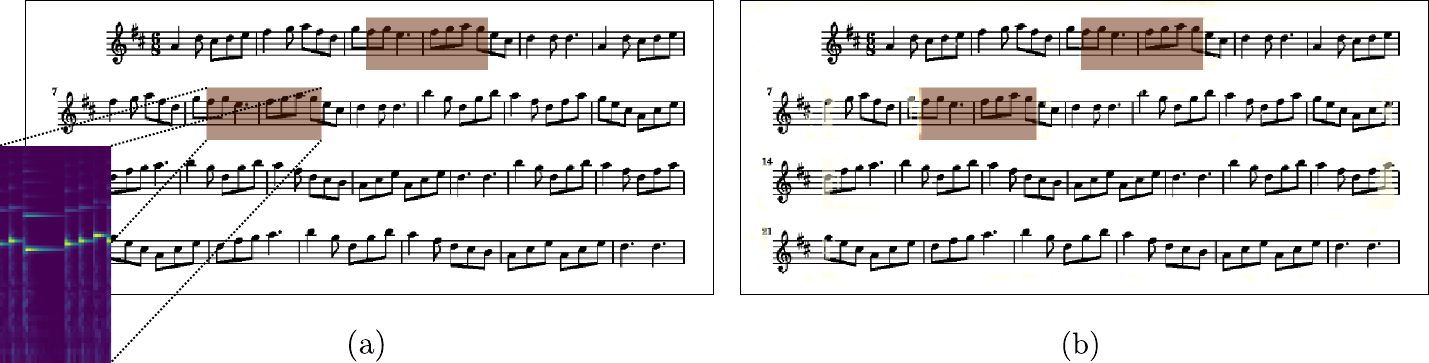}
		\caption{Comparison of (a) a given ground truth that matches with the spectrogram excerpt and (b) the corresponding predictions of the network.}
		\label{fig:gt_pred_comparison}
	\end{figure*}

	To obtain ground truth annotations between the audio and the sheet music, we perform the same automatic notehead alignment as described in \cite{DorferHAFW18_MSMD_TISMIR}. These notehead alignments yield $(x,y)$ coordinate pairs, which are further adjusted such that the $y$ coordinate corresponds to the middle of the staff the respective note belongs to. 
	As we present the network with isolated fixed-size audio (spectrogram) excerpts, \iec we disregard the temporal context, the annotations are no longer unambiguous since an excerpt could match several positions within the sheet image. 
	Thus, we identify all positions in the sheet image that match the audio and create a binary mask with the same shape as the downscaled score page. At positions of a match, this mask contains rectangular regions of height 20 and a width that depends on the distance between the first and last note in the audio excerpt (see \Fig{fig:gt_pred_comparison}).
	The task of the U-Net is to predict a corresponding mask, given a score page and some audio excerpt.
	
	Audio is sampled at 22.05~kHz and processed at a frame rate of 20 frames per second. The DFT is computed for each frame with a window size of 2048 samples and then transformed with a logarithmic filterbank that processes frequencies between 60~Hz and 6~kHz, yielding 78 log-frequency bins. The conditioning network is presented with 40 consecutive frames, or roughly two seconds of audio.
	
	\subsection{Experimental Setup}
	As shown in \Fig{fig:architecture}, we introduce FiLM layers in all U-Net blocks. In our experiments, we test and compare several settings with the conditioning mechanism being activated in different parts of the architecture. As we cannot test all possible combinations due to computational limitations, we choose a subset which we think allows us to assess the influence of the FiLM layers on the final model performance.
	
	As an optimization target we minimize the \emph{Dice} coefficient loss \cite{MilletariNA16_VNet_3DV}, defined as
	
	\begin{equation}
	D(\mathbf{p}, \mathbf{g}) = 1 - \frac{2 \sum_i^N p_i g_i}{\sum_i^N p_i^2 + \sum_i^N g_i^2},
	\label{eq:dice_loss}
	\end{equation}
	where $\mathbf{p}$ and $\mathbf{g}$ are vectors containing the predicted probabilities and ground truth, respectively.
	The advantage of the \emph{Dice} coefficient loss compared to, \egc \emph{binary cross-entropy},  is that it inherently deals with class imbalance. This is important as only a small portion of the sheet image corresponds to a given audio excerpt.
	To optimize the target we use Adam \cite{KingmaB14_Adam_ICLR} with default parameters, an initial learning rate of $0.001$, $L^2$ weight decay with a factor of $5e^{-5}$ and a batch size of $32$. The weights of the network are initialized orthogonally \cite{SaxeMG13_OrthogonalInit_arxiv} and the bias is set to zero. If the loss on the validation set does not decrease for 2 epochs we halve the learn rate. This is repeated five times. The trained model parameters with the lowest validation loss will be used for the final evaluation on the test set.
	
	Additionally, we apply data augmentation on the sheet images by shifting them along the x and y axis. Note that our goal currently is not to generalize to scanned or handwritten scores, which would require more sophisticated augmentation techniques (\egc as shown in \cite{VanDerWelU17_OMRSeq2Seq_arxiv}), but to show the network that note patterns can occur anywhere in the score image. 
	
	\subsection{Results}
	\Table{tab:eval_architecture} reports performance measures for different conditioning scenarios on the test set. They are defined as
		
	\begin{eqnarray*}
		\text{Precision}  =  \frac{tp}{\mathit{tp + fp}}, & & 
		\text{Recall} = \frac{tp}{\mathit{tp + fn}}
	\end{eqnarray*}
	\begin{equation*}
	F_1  =  \frac{2\cdot tp}{2\cdot \mathit{tp + fp + fn}}
	\end{equation*}
	with $\mathit{tp}$ the number of pixels correctly predicted  as 1, $\mathit{fp}$ the pixels falsely predicted as 1 and $\mathit{fn}$ the pixels falsely predicted as 0. The predictions are binarized with a threshold of $0.5$. Note that the optimization target given in \Eq{eq:dice_loss} closely relates to the $F_1$ score, as the originally defined \emph{S{\o}rensen-Dice} coefficient corresponds to the $F_1$ score in the binary case.
	
	Overall, we observe that the performance is high in all tested scenarios, with an $F_1$ score greater than $0.88$. In all cases, the recall is higher than the precision, which could be improved by choosing a higher threshold value than $0.5$. The worst performance results  when the FiLM layer is only activated in the bottleneck and decoding blocks (E-I). We see a similar performance in terms of precision when we apply the conditioning mechanism only in the bottleneck block (E). This suggests that the FiLM layers might be more effective in the encoding part of the network, which is further substantiated by the performance of the conditioning mechanism in encoding blocks (A-E).
	Nevertheless, a marginally higher $F_1$ score is achieved when FiLM layers are applied both during encoding and decoding in blocks (C-G). 
	This indicates that amount and location at which conditioning information is supplied to the feature extraction network needs to be chosen carefully. A change in the overall resolution and depth of the feature extracting U-Net, will likely necessitate a re-tuning of these hyperparameters.
	
	\begin{table}[t]
		\small
		\centering
		\begin{tabular}{lccc}
			\multicolumn{4}{c}{Nottingham (42 test pieces)} \\
			\toprule
			\textbf{Architecture} & \textbf{Precision} & \textbf{Recall}  & $\mathbf{F_1}$ \textbf{Score}\\
			\midrule
			FiLM Layers (E)   & 0.8647 & 0.9216 & 0.8922  \\
			FiLM Layers (D-F) & 0.8785 & 0.9292 & 0.9031  \\
			FiLM Layers (C-G) & 0.8929 & 0.9336 & 0.9128  \\
			FiLM Layers (B-H) & 0.8980 & 0.9261 & 0.9118  \\
			FiLM Layers (A-I) & 0.8903 & 0.9169 & 0.9034  \\
			FiLM Layers (A-E) & 0.8933 & 0.9267 & 0.9097  \\
			FiLM Layers (E-I) & 0.8674 & 0.9104 & 0.8884  \\
			\midrule
			
		\end{tabular}
		\small \caption{Comparison of different FiLM layer combinations. In each scenario, we evaluate the trained model parameters with the lowest validation loss. (D-F) means the FiLM layers in block D, E and F are active.
		}
		\label{tab:eval_architecture}
	\end{table}
	
	\section{Conclusion and Future Work}\label{sec:conclusion}
	We have introduced an architecture capable of inferring corresponding positions for audio excerpts in full score sheet images with high $F_1$ score.
	Although this is not a full score following system yet, we believe this work is a necessary first step towards a fully integrated system that can track musical performances in images of sheet music, without the need for cumbersome preprocessing steps such as OMR.
	The temporal context needed for the last step could come from the hidden state of a recurrent neural network, to be used as temporal conditioning information, in a similar way as described in \cite{PerezSDVDC18_FILM_AAAI}.
	
	Currently, the system has only been tested on monophonic piano music. Using the Multimodal Sheet Music Dataset (MSMD) \cite{DorferHAFW18_MSMD_TISMIR}, this can be further extended to more complex scores with polyphonic music. As the scores in this dataset are often several pages long, one could adapt the proposed architecture to either take multiple pages as channel inputs and predict the final position probabilities for all pages at the same time, or predict the positions for one page at a time and move to the next page once the end of the current one is reached.
	Future work will explore the generalization performance of the system, both in terms of sheet image variations caused by lower quality scans, and differences in musical performances, where tempo, volume and timbre varies.

	\section*{Reproducibility}
	In the interest of reproducible research, we make both code and data available on-line, along with detailed instruction on how to recreate the reported results. \url{https://github.com/CPJKU/audio_conditioned_unet}
	
	\section*{Acknowledgment}
	This project has received funding from the European Research Council (ERC) under the European Union's Horizon 2020 research and innovation program (grant agreement number 670035, project "Con Espressione"). 
	
	\bibliographystyle{ieeetr}
	\bibliography{audio_conditioned_unet}
	
\end{document}